\documentclass[10pt,twocolumn,letterpaper]{article}
\usepackage{multirow}
\usepackage{soul}
\usepackage{cuted}
 \usepackage[pagenumbers]{cvpr} % To force page numbers, e.g. for an arXiv version

\usepackage{algorithm}
\usepackage{algorithmic}
\usepackage{bm}

\usepackage{amssymb}

\usepackage{booktabs} %
\usepackage{tikz}
\usepackage{pgfplots}
\usetikzlibrary{positioning}
\usetikzlibrary{calc}
\usepackage{graphicx}
\usepackage{subcaption}
\usepackage{etoolbox,lipsum}
\usepackage{xcolor}
\usepackage{afterpage}
\definecolor{lightred}{HTML}{F08080}    % 对应“first”
\definecolor{lightorange}{HTML}{FDBA74} % 对应“second”
\definecolor{lightyellow}{HTML}{FDE68A} % 对应“third”
\definecolor{skyblue}{rgb}{0.53, 0.81, 0.98}
\usepackage{colortbl}
\usepackage{etoolbox,lipsum}

\usepackage[utf8]{inputenc}
\usepackage{amsmath}
\usepackage{tikz}
\usetikzlibrary{positioning}
\usetikzlibrary{calc}
\usepackage{caption}

\usepackage{pifont}

\newcommand{\cmark}{\ding{51}}
\newcommand{\xmark}{\ding{55}}

\makeatletter
\DeclareRobustCommand\onedot{\futurelet\@let@token\@onedot}
\def\@onedot{\ifx\@let@token.\else.\null\fi\xspace}

\makeatother

\definecolor{cvprblue}{rgb}{0.21,0.49,0.74}
\usepackage[pagebackref,breaklinks,colorlinks,allcolors=cvprblue]{hyperref}

\def\paperID{***} % *** Enter the Paper ID here
\def\confName{CVPR}
\def\confYear{2026}

\title{Prune Wisely, Reconstruct Sharply: Compact 3D Gaussian Splatting via Adaptive Pruning and Difference-of-Gaussian Primitives}
\author{
  Haoran Wang,
  Guoxi Huang,
  Fan Zhang, 
  David Bull, and
  Nantheera Anantrasirichai\\
  School of Computer Science,
  University of Bristol,
  Bristol, UK \\\tt\small{\{yp22378,guoxi.huang,fan.zhang,dave.bull,n.anantrasirichai\}@bristol.ac.uk}}

\begin{document}
\maketitle

\begin{strip}
    \centering
    \vspace{-1.0cm}
    \includegraphics[width=1\textwidth]{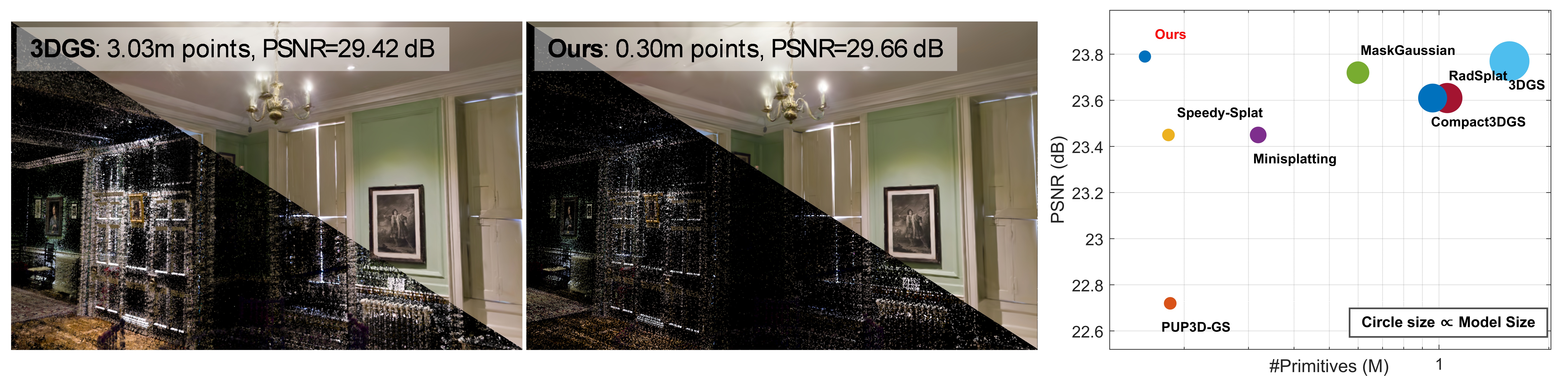}
    \vspace{-0.2cm}
    \captionof{figure}{Performance improvement of the proposed pruning strategy and 3D Difference-of-Gaussian (DoG) primitives. Comparison on the Drjohnson scene, where the original 3DGS method is shown on the left and our result in the middle panel. The right panel shows PSNR versus the number of primitives on the Tanks and Temples dataset~\cite{knapitsch2017tanks} across various 3DGS-based methods.}
    %\captionof{figure}{We proposed a new Gaussian Splatting pruning pipeline. One example from the \textbf{drjohnson} scene is shown in the left figure. Compared with the original 3DGS method, our method can retain a very sparse point cloud while achieving excellent performance. In the Right Figure, we show that our method achieves the highest PSNR with the fewest primitives on the Tanks and Temples~\cite{knapitsch2017tanks} dataset. The size of the point represents the model complexity.}
    \label{fig:teaser}
    %\vspace{-0.2cm}
\end{strip}

\begin{abstract}
Recent significant advances in 3D scene representation have been driven by 3D Gaussian Splatting (3DGS), which has enabled real-time rendering with photorealistic quality. 3DGS often requires a large number of primitives to achieve high fidelity, leading to redundant representations and high resource consumption, thereby limiting its scalability for complex or large-scale scenes. Consequently, effective pruning strategies and more expressive primitives that can reduce redundancy while preserving visual quality are crucial for practical deployment. We propose an efficient, integrated reconstruction-aware pruning strategy that adaptively determines pruning timing and refining intervals based on reconstruction quality, thus reducing model size while enhancing rendering quality. Moreover, we introduce a 3D Difference-of-Gaussians primitive that jointly models both positive and negative densities in a single primitive, improving the expressiveness of Gaussians under compact configurations. Our method significantly improves model compactness, achieving up to 90\% reduction in Gaussian-count while delivering visual quality that is similar to, or in some cases better than, that produced by state-of-the-art methods. 
Code will be made publicly available.
\end{abstract}    
\section{Introduction}
\label{sec:intro}

Novel View Synthesis (NVS) aims to generate photorealistic images from an unseen viewpoint based on a limited set of input observations. This research topic was initially dominated by NeRF-based methods \cite{mildenhall2021nerf, ma2022deblur, wang2021nerf, deng2022depth, fridovich2022plenoxels}, which implicitly represent a scene as a continuous radiance field and estimate volume density and colors using neural networks. While these methods have attracted significant attention and supported a wide range of applications~\cite{anantrasirichai2025artificial}, they commonly suffer from computational overhead and render very slowly, making them unsuitable for real-time applications.

3D Gaussian Splatting (3DGS) \cite{kerbl20233d} has recently emerged as an alternative efficient and high-quality representation for 3D scene reconstruction and novel view synthesis. By explicitly modeling a scene as a set of spatially distributed Gaussian primitives, 3DGS achieves real-time photorealistic rendering while maintaining differentiability for optimization. Due to these advantages, 3DGS has rapidly become the dominant approach in 3D reconstruction. Nonetheless, achieving high-quality reconstruction typically requires a large number of Gaussian primitives, many of which are redundant or only contribute marginally to the final render~\cite{fang2024mini2}. This redundancy results in unnecessary memory consumption and computational burden, significantly limiting scalability for large or complex scenes. 

To improve the efficiency of 3DGS, several pruning approaches have been proposed \cite{fang2024mini, hanson2025pup, fan2024lightgaussian}. Although these have achieved promising results, most of them prune at fixed training iterations and use uniform refinement intervals, hence disregarding the dynamic nature of the reconstruction process. Such inflexible schedules often lead to unstable optimization. Early pruning may remove necessary primitives, while late pruning usually provides little efficiency gain. To address these issues, we propose a reconstruction-aware pruning framework that adaptively determines when to prune based on reconstruction quality. Instead of pruning in predetermined iterations, our method analyzes reconstruction quality and automatically identifies the optimal pruning time, adjusting the pruning ratio as the procedure progresses. This adaptive mechanism enables the model to maintain stability during training and progressively achieve compact representations.  To retain reconstruction quality after compression, we also propose a Spatio-spectral Pruning Score that measures the importance of a Gaussian primitive in both spatial and spectral domains.

A second key factor that limits the performance of compact models lies in the nature of 3D Gaussian primitives. It is difficult for a small number of smooth Gaussian kernels to capture fine details accurately. Consequently, we introduce a 3D Difference-of-Gaussians (3D-DoG) primitive, a novel variant of the standard Gaussian that models both positive and negative spatial responses. Its positive-density lobe still contributes to rendering in the same way as normal 3D Gaussians, while its negative-density component can implement color subtraction in $\alpha$-blending. Compared to the original primitive, the 3D-DoG is more responsive to fine geometric details and edges, thereby preserving sharper structures under compact configurations. \autoref{fig:teaser} shows significant improvements with the proposed framework.

Our main contributions are summarized as follows:
\begin{enumerate}[label=\arabic*), leftmargin=13pt, nolistsep]
\item Reconstruction-aware Pruning Scheduler (RPS). We propose a novel dynamic pruning strategy to address the inefficiency and instability of existing pruning schedules, where fixed or uniform pruning may remove important primitives too early or occur too late to yield meaningful efficiency gains.% that adaptively determines when to prune based on reconstruction progress, enabling early compactness and stable convergence.

\item Spatio-spectral Pruning Score (SPS). We design a new importance ranking mechanism that incorporates spectral information into the importance ranking of Gaussian primitives; it enables stable pruning, where iteratively removed low-score points are relatively unimportant across both spatial and spectral domains.

\item 3D Difference-of-Gaussians (3D-DoG) primitive. We introduce a new Gaussian variant with both positive and negative components, functioning as a primitive with intrinsic contrast to capture fine structural details.

\item Scalable efficiency. Our approach achieves comparable or superior rendering quality with 90\% fewer primitives on the Mip-NeRF 360, Tanks \& Temples, and Deep Blending datasets, offering a more scalable and efficient 3DGS framework for complex scenes.

\end{enumerate}
%-------------------------------------------------------------------------

\section{Related Work}
\label{sec:formatting}
\subsection{3D Gaussian Splatting}
3D Gaussian Splatting (3DGS) ~\cite{kerbl20233d} is an explicit representation method that uses a cloud of differentiable 3D Gaussians to model a scene. Its popularity is  due to its ability to render in real-time with high-fidelity reconstruction. In recent years, several works have extended its capabilities in quality, robustness, and generality. To address aliasing artifacts arising from unconstrained 3D frequencies, Mip-Splatting ~\cite{yu2024mip} and multi-scale splatting methods ~\cite{yan2024multi} introduced 3D smoothing and mipmap-based level-of-detail rendering. Deblurring variants~\cite{chen2024deblur, lee2024deblurring} improved rendering from blurred or defocused inputs by modeling motion trajectories and defocus effects. Beyond static scenes, dynamic extensions such as 4DGS and Gaussian-Flow~\cite{Wu_2024_CVPR, lin2024gaussian, huang2024sc} incorporated temporal deformation networks to represent 4D scenes, allowing for moderate motion capture. Further, frequency regularization ~\cite{zhang2024fregs} and pixel-aware density control ~\cite{zhang2024pixel} help maintain detail consistency and temporal stability across scales and viewpoints. More recently, Liberated-GS ~\cite{pan2025liberated} decoupled 3DGS from traditional Structure-from-Motion (SfM)~\cite{westoby2012structure} point clouds, achieving self-contained reconstruction without external initialization. EnliveningGS ~\cite{shen2025enliveninggs} explored the active locomotion of Gaussian primitives, extending 3DGS towards physically plausible and dynamic behaviors. In parallel, FlashGS ~\cite{feng2025flashgs} improved computational efficiency and scalability for large-scale and high-resolution rendering, making 3DGS more practical for real-world deployment.

\begin{figure*}[ht]
  \centering
    \vspace{-0.3cm}
   \includegraphics[width=1\linewidth]{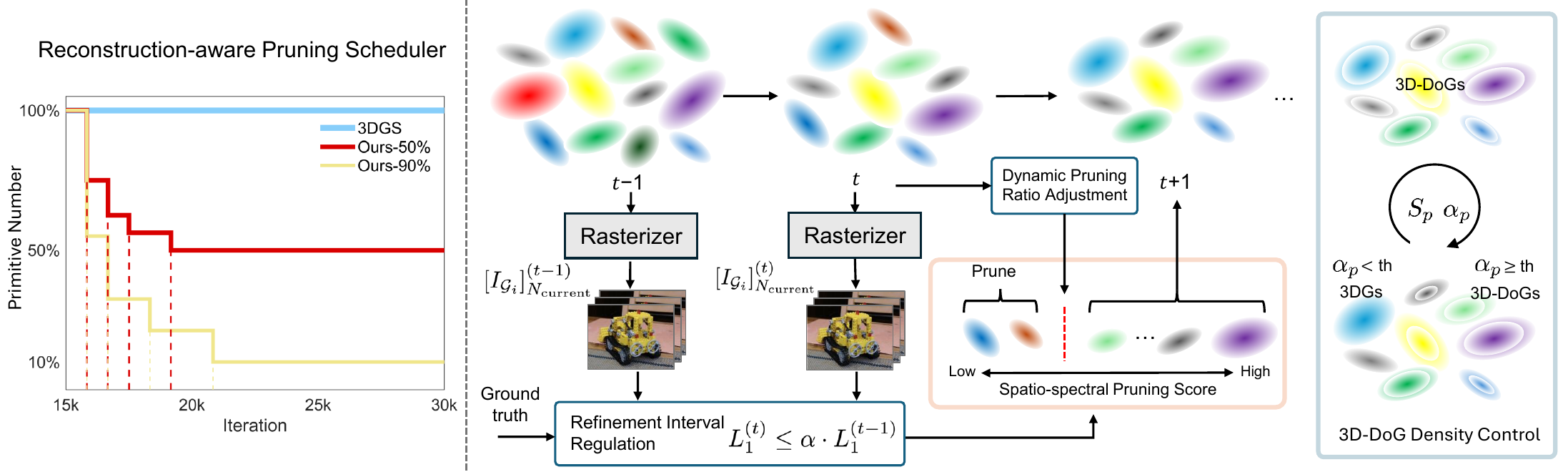}

   \caption{(Left) Gaussian primitive count comparison. Our method adaptively adjusts the refinement settings to meet different pruning targets, such as the 50\% and 90\% pruning ratios shown in the figure. (Right) Overview of the Reconstruction-aware Pruning Scheduler and 3D-DoG Density Control. We use L1 loss as a reconstruction quality indicator to dynamically determine pruning timing and ratio throughout optimization. In addition, we activate 3D-DoG after pruning and adaptively control its density.}
   \label{fig:workflow}
         \vspace{-3mm}
\end{figure*}

\subsection{Pruning Gaussian Splats}

Although the aforementioned works have substantially advanced 3D Gaussian Splatting (3DGS), improving its computational efficiency remains a critical challenge. The naive density control strategy employed often introduces redundancy, thereby increasing computational overhead~\cite{bagdasarian20253dgs}. Pruning unnecessary points has thus emerged as the most common solution to this problem. Typically, Gaussians are ranked by importance, with the least important discarded. Several works~\cite{navaneet2024compgs, zhang2025gaussianspa} approximate Gaussian importance using opacity values. Later studies proposed dedicated pruning scores: RadSplat~\cite{niemeyer2025radsplat} defines an efficient score based on the accumulated ray contributions of individual Gaussians; Mini-Splatting~\cite{fang2024mini} aggregates blending weights to form a pruning score; PuP-3DGS~\cite{hanson2025pup} evaluates spatial sensitivity; and Speedy-Splat~\cite{hanson2025speedy} further leverages per-Gaussian gradients for improved performance. Other works, such as MaskGaussian~\cite{liu2025maskgaussian} and LP-3DGS~\cite{zhang2024lp}, employ Gumbel-Softmax~\cite{jang2016categorical} to learn adaptive masks for Gaussian importance estimation.

Unlike previous heuristic pruning methods, our approach performs adaptive, reconstruction-aware pruning, leading to more efficient compression without sacrificing quality.

\section{Method}

The diagram of our method is illustrated in~\autoref{fig:workflow}. We propose a Reconstruction-aware Pruning Scheduler to assist the progressive prune-refine process. With this adaptive pruning strategy, we can efficiently compress the 3DGS model at reasonable ratios and appropriate time steps. Subsequently, our method introduces a divergent primitive model, 3D-DoG (3D Difference-of-Gaussians), to compensate for detail loss caused by pruning. Consequently, the obtained compact 3DGS model can retain rendering quality and geometric fidelity with enhanced efficiency.
\subsection{Preliminaries}
\label{3dgsPreliminaries}
3DGS~\cite{kerbl20233d} models a scene as a collection of 3D Gaussian primitives. Each Gaussian primitive $\mathcal{G}_i$ is defined by a center 3D position $\mu_i$, a covariance matrix $\Sigma_i$, an opacity $\alpha_i$, and a view-dependent colour $c_i$ defined by using spherical harmonics (SH). To facilitate a differentiable optimization, the covariance matrix $\Sigma_i$ is factorized into a rotation matrix $R_i \in \mathbb{R}^{3 \times 3}$ and a scaling matrix $S_i \in \mathbb{R}^{3 \times 3}$ in the 3D Gaussian primitive.
\begin{equation}
\Sigma_i = R_i S_i S_i^{T} R_i^{T}.
\end{equation}
To render images, 3D Gaussians need to be projected from 3D to 2D. Each 3D Gaussian undergoes a viewing transformation $W$ that projects it onto the image plane, and its 2D covariance matrix can be calculated as follows:
\begin{equation}
\Sigma'_i = J W \Sigma_i W^{T} J^{T},
\end{equation}
where $J$ denotes the Jacobian matrix~\cite{waldron1985study} corresponding to an affine approximation of the projective transformation.

The original 3D Gaussian primitive can be considered as a smooth kernel; its density reaches a peak at its center and gradually declines to 0 as the rendered pixel moves away from its center. As a result, the effective blending weight $\alpha'_i$ of a 3D Gaussian $G_i$ in pixel $x$ can be computed as:
\begin{equation}
\label{alphai}
    \alpha'_i = \alpha_i \exp\!\left(-\tfrac{1}{2} (x - \mu'_i)^{\!T} (\Sigma'_i)^{-1} (x - \mu'_i)\right),
\end{equation}
where $\mu'_i$ refers to the 2D Gaussian’s mean after projecting 3D coordinate $\mu_i$
The pixel colour formation following the $\alpha$-blending scheme as follows:
\begin{equation}
C = \sum_{i} c_i\, \alpha'_i \prod_{j=1}^{i-1} (1 - \alpha'_j),
\end{equation}
$c_i$ corresponds to the colour decoded from the coefficients of the 3D Gaussian spherical harmonics $i$-th, which can vary depending on the viewing direction.

During 3DGS optimization, the 2D position gradients of 3D Gaussians serve as a key factor in controlling the model size growth \cite{wang2025uw}. However, this densification mechanism leads to over-densification, resulting in a high proportion of redundant Gaussian points \cite{huang2025decomposing}. Consequently, an additional pruning process is needed to achieve a trade-off between performance and efficiency. In addition, the smoothness of the 3D Gaussian primitive limits its ability to represent fine details in a compact model. Therefore, a more expressive primitive design is highly desirable. 

\subsection{Reconstruction-aware Pruning Scheduler}
\label{ssec:RApruning}
Previous work has typically  performed pruning at fixed iterations and with a preset ratio~\cite{hanson2025speedy}. However, this naive strategy suffers from two significant limitations. First, the difficulty of reconstruction varies significantly across scenes, yet conventional iterative pruning methods adopt fixed pruning intervals regardless of this variation. In rapidly converging scenes, overly long intervals result in redundant points during rendering, wasting computational resources. Conversely, in more complex scenes, pruning too frequently prevents the point cloud from fully training, resulting in sub-optimal performance and loss of fine details. Second, pruning with fixed ratios neglects the changing redundancy level of the point cloud. As pruning progresses, redundancy naturally decreases; maintaining a large ratio then leads to over-pruning in later stages, thereby degrading reconstruction quality. 

To address these issues, we propose a Reconstruction-aware Pruning Scheduler (RPS) that dynamically adjusts the pruning interval based on the reconstruction loss and progressively reduces the number of pruned primitives at each step. In this way, RPS enables more adaptive, efficient, and detail-preserving pruning across diverse scenes. An overview of the RPS process is shown in~\autoref{fig:workflow}, and its details are described below.

\begin{algorithm}[t]
\caption{Refinement Interval Regulation strategy}
\label{alg:prune_refine}
\resizebox{\columnwidth}{!}{%
\begin{minipage}{\columnwidth}
\begin{algorithmic}[1]
\REQUIRE Current Model $\mathcal{M}_t$, reconstruction losses $L_1^{(t)}$, $L_1^{(t-1)}$, threshold $\beta$, and maximum interval $Iter_{\max}$.
\IF{$L_{1}^{(t)} \le \beta \cdot L_{1}^{(t-1)}$}
    \STATE $\mathcal{M}_{t+1} = \operatorname{\textcolor{red}{prune}}(\mathcal{M}_t)$
\ELSE
    \STATE $\operatorname{\textcolor{red}{refine}}(\mathcal{M}_t)$ until criterion met or $Iter_{\max}$ reached
\ENDIF
\end{algorithmic}
\end{minipage}%
}

\end{algorithm}

\vspace{2mm}\noindent\textbf{Refinement Interval Regulation.}
Previous refinement strategies~\cite{fang2024mini} have pruned the 3D point cloud after each refinement step. However, the question of when to prune (i.e. how long the refining interval should be) remains underexplored. We address this by adaptively modulating the refinement interval throughout the pruning process. After each pruning step, we use the average L1 loss as a criterion to determine whether a further pruning operation should be executed. Let $L_{1}^{(t)}$ denote the L1 reconstruction loss computed on the whole training dataset after the $t$-th pruning step, and $L_{1}^{(t-1)}$ denote the loss recorded before the $t$-th pruning step. To integrate the prune–refine procedure into the training process rather than applying it as a post-processing step~\cite{fan2024lightgaussian, hanson2025pup}. We set a pruning criterion as follows:
\begin{equation}
    L_{1}^{(t)} \le \beta \cdot L_{1}^{(t-1)},
    \label{eq:prune_condition}
\end{equation}
where we set $\beta=0.95$. If the above condition is satisfied, this indicates that the reconstruction quality has improved. In this case, we proceed to the next pruning step. Otherwise, the structure remains unchanged and refinement continues until the condition in~\eqref{eq:prune_condition} is met or the maximum refinement interval $Iter_{\max}$, which is set to 2000 iterations, is reached to avoid a bottleneck. The process is summarized in Algorithm~\autoref{alg:prune_refine}. This procedure is carried out every 500 iterations until the pruning target is reached, at the cost of a slight degradation in efficiency.

\vspace{2mm}\noindent\textbf{Dynamic Pruning Ratio Adjustment.}
We observe that as the overall 3DGS model size decreases, pruning increasingly degrades the reconstruction quality, which is in agreement with the findings reported in~\cite{ali2025compression}. A key reason for this is that the number of redundant GS primitives varies during the pruning process. When redundancy is high, many 3D points overlap spatially and contribute little to the final rendering; aggressive pruning is desirable. However, as the model size decreases, most remaining Gaussians are crucial for preserving fine geometric and photometric details. Building on this observation, we introduce a novel dynamic pruning scheduler in which the removal rate decreases over time. Aggressive pruning is allowed in the early stages, and only a relatively few points are removed in the later stages. Moreover, we integrate multistep pruning into the 3DGS model optimization rather than running them sequentially \cite{fan2024lightgaussian}, thereby improving training efficiency and enabling the model to adaptively balance performance and sparsity. Inspired by \cite{chen2025dashgaussian}, we control the pruning ratio of the round $t$ as follows:
\begin{equation}
    N^{(t)} =
    N_{\text{current}}
    -
    \bigl( N_{0} - N_{\text{target}} \bigr)
    \cdot
    \frac{1}{2^t},
\end{equation}
\begin{equation}
    R^{(t)} = \frac{\bigl( N_{\text{current}} - N^{(t)} \bigr)}{\bigl(N_{\text{current}}\bigr)},
\end{equation}
where $N^{(t)}$ and $R^{(t)}$ denote the desired number of Gaussians and computed pruning ratio of pruning round $t$, $N_{0}$ is the number of Gaussians after complete densification, $N_{\text{target}}$ is the preset target number. We use the current number of primitives $N_{\text{current}}$ and $N^{(t)}$ to obtain the pruning ratio for the current round. By dynamically adjusting the pruning ratio, the model size can be significantly reduced in the early iterations, and the training process further accelerated.

\vspace{2mm}\noindent\textbf{Spatio-spectral Pruning Score.}
To further enhance the effectiveness of pruning, we introduce a new criterion for evaluating the importance of each Gaussian. Specifically, we develop a Spatio-spectral Pruning Score, which measures the contribution of each primitive from both spatial and spectral perspectives. Previous work~\cite{hanson2025speedy} employs a pruning score that enables a fast estimation, expressed as follows.
\begin{equation}
    \tilde{U}_i = \left( \nabla_{g_i} I_{\mathcal{G}} \right)^2,
\end{equation}
where $\tilde{U}_i$ denotes the efficient pruning score of the $i$-th Gaussian, $\nabla_{g_i}$ denotes the gradient of the parameters of $i$-th Gaussian with respect to the rendered image $I_{\mathcal{G}}$.

However, this importance evaluation concentrates on the spatial domain but ignores the frequency domain. Incorporating the frequency domain ensures that Gaussians essential for preserving sharp structures are not mistakenly pruned. Hence, we propose a spectral pruning score:
\begin{equation}
\tilde{U}_i^{f}
=
\sum_{\omega \in \Omega}
w(\omega)\,
\bigl\lvert \nabla_{g_i}\,\hat{I}_{\mathcal{G}}(\omega) \bigr\rvert^{2},
\quad
\hat{I}_{\mathcal{G}} = \mathrm{FFT}\!\left( I_{\mathcal{G}} \right),
\end{equation}
where $\lvert\cdot\rvert$ is the complex modulus and $w(\omega)\!\ge\!0$ is a frequency weight that emphasizes informative bands. We use a radial schedule:
\begin{equation}
w(\omega)=\left(\frac{\|\omega\|}{\omega_{\max}}\right)^{\gamma_f},\ {\gamma_f}>0,\quad
w(\mathbf{0})=0,
\end{equation}
By combining a spatially aware and spectrally aware importance score $\tilde{U}_i$ and $\tilde{U}_i^{f}$, we obtain our Spatio-spectral Pruning Score (SPS) $\tilde{U}_i^{*}$:
\begin{equation}
    \tilde{U}_i^{*}
    =
    \lambda_s \cdot
    \frac{ \left( \nabla_{g_i} I_{\mathcal{G}} \right)^{2} }
         { \| \tilde{U} \|_2 }
    +
    \lambda_f \cdot
    \frac{ \left( \nabla_{g_i}\, \mathrm{FFT}( I_{\mathcal{G}} ) \right)^{2} }
         { \| \tilde{U}^{f} \|_2 } ,
\end{equation}
where $\lambda_s$ and $\lambda_f$ are weighting coefficients that balance the contributions of the spatial- and frequency-domain gradient terms, respectively. This formulation allows the model to consider both spatial variations jointly and frequency-domain characteristics for more stable pruning.

\subsection{3D Difference-of-Gaussians (3D-DoG)}
\label{3DDOG}
\begin{figure}
  \centering

   \includegraphics[width=1.0\linewidth]{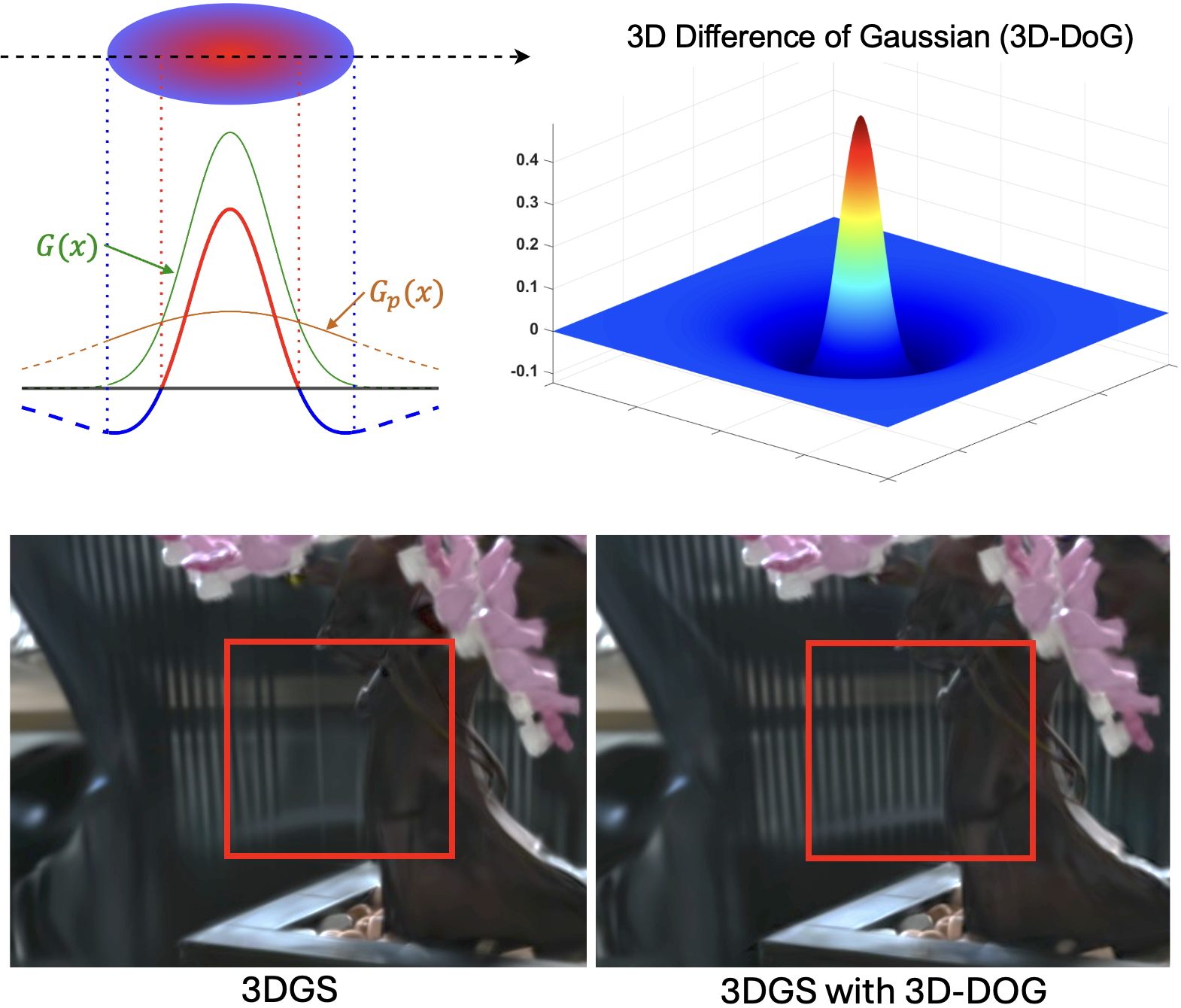}

   \caption{(Top) Illustration of the proposed 3D-DoG primitive in 1D and 3D, featuring a positive-density peak and a negative-density ring. (Bottom) 3DGS with 3D-DoG primitives achieves better detail representation.}
   \label{fig:3D-DoG}
      \vspace{-4mm}
\end{figure}

\noindent\textbf{3D-DoG Kernel.}
   After pruning, we found that the compact model suffers a very significant loss in details. This issue is due to the intrinsic properties of 3DGS: splatting operates only in the positive-density domain. To represent boundary or texture areas, vanilla 3DGS can only stack dozens of small, overlapping Gaussians of different shapes to capture fine details~\cite{shi2025sketch}. However, after aggressive pruning, there are not enough primitives to reconstruct those areas. To address this, we introduce a novel 3D Difference-of-Gaussian (3D-DoG) primitive, which can be viewed as the difference between a primary-Gaussian and a pseudo-Gaussian as shown in \autoref{fig:3D-DoG} (top).

We define the 3D-DoG primitive as follows:
\begin{equation}
  DoG(x) = G(x) - G_{p}(x),
  \label{eq:important}
\end{equation}
where $G_{p}(x)$ is a pseudo-Gaussian that shares the same center coordinates of the kernel and all other parameters with the primary-Gaussian $G(x)$ except for opacity and scales. Consequently, compared to the vanilla 3D Gaussians, 3D-DoG introduces four additional learnable parameters to define the pseudo-Gaussian, which are the opacity factor $f^{\alpha}$ and the scaling factors $[f^{s}_{x}, f^{s}_{y}, f^{s}_{z}]$. The scalings $S_{p}$ and opacity $\alpha_{p}$ are defined as follows:
\begin{equation}
S_{p} = \begin{bmatrix}
s_x & 0 & 0 \\
0 & s_y & 0 \\
0 & 0 & s_z 
\end{bmatrix} \cdot \begin{bmatrix}
f_x & 0 & 0 \\
0 & f_y & 0 \\
0 & 0 & f_z 
\end{bmatrix}^{T},\\
\end{equation}
\begin{equation}
    \alpha_{p} = f^{\alpha} \cdot \alpha,
\end{equation}
where $[s_x, s_y, s_z]$ and $\alpha$ are the scaling parameters and the opacity of the primary Gaussian, respectively. Combined with other existing parameters, all are set to values less than 1.0 to ensure that our 3D-DoG profile maintains a positive lobe that governs the radiance, consistent with the original 3D Gaussian formulation. The surrounding negative-density ring serves as a contrast component, subtracting the color of neighboring pixels it overlaps with~\cite{zhu20253d}. In other words, the 3D-DoG inherently encodes contrast within its construction. Compared to 3D Gaussians, which capture overall global geometry, 3D-DoGs have a sharper response and can thus enhance features such as boundaries when the number of primitives is limited. This advantage is clearly demonstrated in \autoref{fig:3D-DoG} (bottom).

\begin{figure*}[!t]
  \centering
\vspace{-5mm}
   \includegraphics[width=1.0\linewidth]{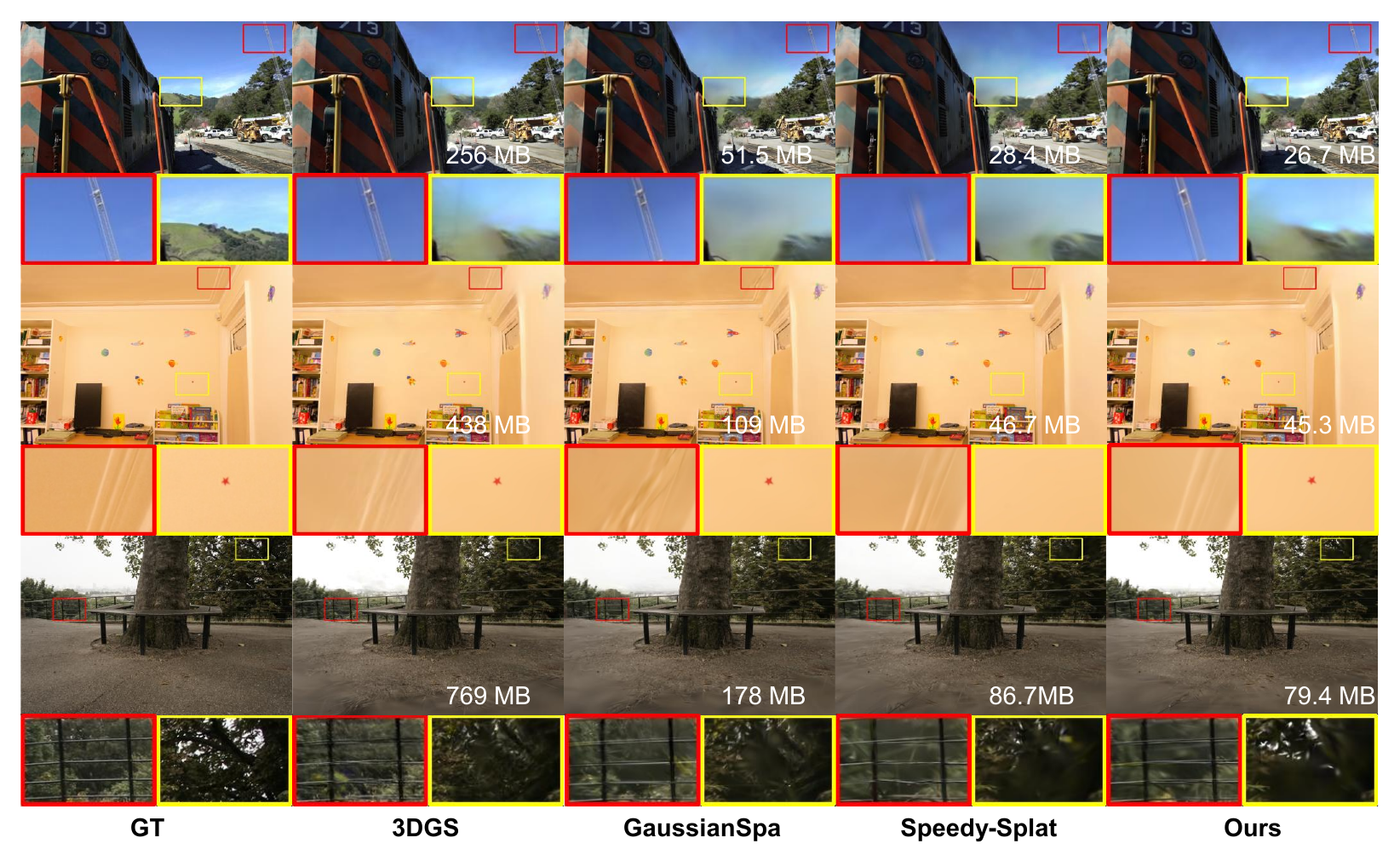}
    \vspace{-5mm}
    \caption{\textbf{Novel view rendering comparison with the baselines.} Top: \textbf{Train} from the Tanks \& Temples. Middle: \textbf{Playroom} from the Deep Blending dataset. Bottom: \textbf{Treehill} from the Mip-NeRF 360 dataset. We have shown details below the images. Best viewed when zoomed in.}
   \label{fig:Result}
   \vspace{-4mm}
\end{figure*}
\vspace{2mm}\noindent\textbf{3D-DoG Density Control.}
Since 3D-DoG introduces additional computational overhead, our observations indicate it takes longer to cover a dense point cloud. Thus, we only introduce 3D-DoG primitives in the compact model.

After the pruning phase, we employ 3D-DoG primitives to replace all primitives in the compact version model by activating $[f^{s}_{x}, f^{s}_{y}, f^{s}_{z}]$ and $\alpha_{p}$ to allow the adjustment of pseudo-Gaussians, and continue training with the same loss functions. 
By evaluating $\alpha_{p}$ across all primitives, we can identify those with the lowest $\alpha_{p}$ values, indicating that the pseudo-Gaussian representations of these 3D-DoGs have a marginal impact on representations of these primitives. When $\alpha_{p}$ = 0, 3D-DoGs become equivalent to a standard 3D Gaussians. Consequently, 3D-DoGs with $\alpha_{p}$ values below a predefined threshold are degenerated into 3D Gaussian components at each training iteration. Therefore, the computational overhead is reduced with minimal loss of fidelity.

This density control mechanism enables us to adaptively adjust the ratio of different types of primitives in the hybrid representation to accommodate various scene characteristics. By iteratively controlling the 3D-DoG density, our pipeline achieves a balance between efficiency and reconstruction quality. 3D-DoG and 3D Gaussian primitives are jointly optimized to play complementary roles, allowing the resulting mixture model to maintain stability in smooth regions while better recovering fine details.  

\begin{table*}[t]
    \centering
    %\footnotesize
    \setlength\tabcolsep{2pt}
    \captionsetup{singlelinecheck=false}
        \caption{Quantitative evaluation of our method compared to previous work, computed over three datasets: Mip-NeRF 360, Deep Blending and Tanks and Temples. $\uparrow$ refers larger values are better while $\downarrow$ is opposite. The \colorbox{lightred}{best}, \colorbox{lightorange}{second best}, and \colorbox{lightyellow}{third best} results are highlighted. $^{\dagger}$ denotes accelerated diff-gaussian-rasterization module is adopted}
    \resizebox{\textwidth}{!}{
    \begin{tabular} {r | rrrrr | rrrrr | rrrrr}
        \toprule
        \multirow{2}{*}{Method} 
        & \multicolumn{5}{c}{Mip-NeRF 360~\cite{barron2022mip}} & \multicolumn{5}{c}{Deep Blending~\cite{hedman2018deep}} & \multicolumn{5}{c}{Tanks and Temples~\cite{knapitsch2017tanks}} \\
        \cmidrule(l{2pt}r{2pt}){2-6} \cmidrule(l{2pt}r{2pt}){7-11} \cmidrule(l{2pt}r{2pt}){12-16}
        & $\mathrm{ Size}\downarrow$ & PSNR$\uparrow$ & SSIM$\uparrow$ & LPIPS$\downarrow$ &Time$\downarrow$ 
        & $\mathrm{ Size}\downarrow$ & PSNR$\uparrow$ & SSIM$\uparrow$ & LPIPS$\downarrow$ &Time$\downarrow$ 
        & $\mathrm{ Size}\downarrow$ & PSNR$\uparrow$ & SSIM$\uparrow$ & LPIPS$\downarrow$ &Time$\downarrow$ \\
        \midrule
        3DGS$^{\dagger}$~\cite{kerbl20233d}
                          &          645.2 &          27.47 &          0.826 &          0.201 &          17m1s 
                          &          592.7 &          29.75 &          0.904 &          0.244 &          13m51s 
                          &          381.0 &          23.77 &          0.847 &          0.177 &          17m50s \\

        \midrule
        MaskGaussian~\cite{liu2025maskgaussian}
                            &          280.7 &          \cellcolor{lightred}27.43 &          \cellcolor{lightred}0.811 &          \cellcolor{lightred} 0.227  &          \cellcolor{lightyellow}24m11s 
                          &          172.4 &          \cellcolor{lightorange} 29.69  &          \cellcolor{lightred}0.907 &          \cellcolor{lightred}0.244 &          \cellcolor{lightyellow}15m10s 
                          &          140.0 &           \cellcolor{lightorange}23.72 &          \cellcolor{lightred}0.847 &          \cellcolor{lightred} 0.181 &          \cellcolor{lightyellow}13m45s\\
         GaussianSpa~\cite{zhang2025gaussianspa}
                            &          157.4 &        \cellcolor{lightorange}27.26 &          \cellcolor{lightorange}0.807 &          \cellcolor{lightorange}0.239&          24m45s 
                          &          132.5 &         \cellcolor{lightyellow}29.66  &          \cellcolor{lightorange}0.905 &          \cellcolor{lightorange}0.250&          \cellcolor{lightorange}13m49s
                          &          87.3 &          \cellcolor{lightyellow} 23.67&         \cellcolor{lightorange}0.841 &        \cellcolor{lightorange}0.198&          23m11s  \\
        PuP-3DGS~\cite{hanson2025pup}
                          &          \cellcolor{lightyellow}90.6 &           26.67 &          0.786 &         \cellcolor{lightyellow}0.271 &           - 
                          &          \cellcolor{lightyellow}69.9 &           28.85 &          0.881 &        0.302 &           - 
                          &          \cellcolor{lightyellow}43.4 &           22.72 &          0.801 &         0.244 &           - \\
        Speedy-Splat~\cite{hanson2025speedy}
                          &          \cellcolor{lightorange}73.9 &           26.84 &          0.782 &          0.296 &           \cellcolor{lightorange}16m30s 
                          &          \cellcolor{lightred}58.6 &           29.42 &          0.887 &          0.311 &           15m30s
                          &          \cellcolor{lightorange} 43.0  &           23.43 &          0.821 &          0.241 &           \cellcolor{lightorange}9m37s \\
        \midrule
        Ours$^{\dagger}$
                          &          \cellcolor{lightred}65.3 &         \cellcolor{lightyellow} 27.16 &          \cellcolor{lightyellow}0.789 &          0.285 &          \cellcolor{lightred}13m48s
                          &          \cellcolor{lightorange}59.9 &          \cellcolor{lightred}29.87 &          \cellcolor{lightyellow}0.904 &          \cellcolor{lightyellow}0.254 &           \cellcolor{lightred}10m19s 
                          &          \cellcolor{lightred}38.4 &          \cellcolor{lightred}23.79 &          \cellcolor{lightyellow}0.823 &         \cellcolor{lightyellow}0.229 &           \cellcolor{lightred}8m9s \\
        \bottomrule
    \end{tabular}
    }
    \label{tab:main-result}
\end{table*}
\section{Experiments}
\label{sec:Experiment}

\subsection{Experimental Setup}   

\noindent\textbf{Datasets.}
To comprehensively evaluate our method, we use the Mip-NeRf 360~\cite{barron2022mip} dataset, which comprises four indoor and five outdoor scenes. Furthermore, we also tested Deep Blending~\cite{hedman2018deep} and Tanks \& Temples~\cite{knapitsch2017tanks}, which provide two additional indoor and two additional outdoor scenes, respectively. 

\vspace{2mm}\noindent\textbf{Implementation.}
The official code for the 3DGS implementation~\cite{kerbl20233d} is adopted as the backbone. In all experiments, we set a pruning target of 90\%, retaining only 10\% of primitives compared to the original model, since this setting is particularly challenging.

Only the process after the first 15,000 iterations is modified, once the densification has finished. Our Reconstruction-aware Pruning Scheduler is integrated with the rest of the optimization to enable the progressive pruning-refine process until the model size reaches the preset pruning target. The duration of this step is flexibly adjusted as described in \autoref{ssec:RApruning}, but will conclude after no more than 25,000 iterations. After the pruning process finishes, we introduce the 3D-DoG primitives into the model and optimize their overall proportion in the point cloud. Our experiments were carried out on a single RTX 3090 GPU. We adopt the accelerated rasterization CUDA module~\cite{mallick2024taming} to improve efficiency. The forward and backward branches have been modified to adapt to 3D-DoG splatting. More details are available in the supplementary material (SM).

\vspace{2mm}\noindent\textbf{Baseline and Metrics.}
Apart from the original 3DGS\cite{kerbl20233d}, we selected MaskGaussian~\cite{liu2025maskgaussian}, GaussianSpa~\cite{zhang2025gaussianspa}, PuP-3DGS~\cite{mallick2024taming}, and Speedy-Splat~\cite{hanson2025speedy} as baselines; all of these methods aim to compress the original 3DGS model by pruning. MaskGaussian does not specify a pruning ratio, whereas GaussianSpa uses an 80\% pruning ratio when built upon 3DGS. PuP-3DGS and Speedy-Splat both target a 90\% pruning ratio. By testing these representative methods as baselines, we evaluate our approach not only on reconstruction quality but also on memory usage and training speed. We adopt common PSNR, SSIM, and perceptual-based LPIPS to measure the reconstruction quality, and use model size and training time to evaluate computational efficiency. 

\subsection{Results and Discussion}
\noindent\textbf{Quantitative comparison.}
    \autoref{tab:main-result} shows the evaluation results of our method compared with all the benchmarks. Our method achieves state-of-the-art reconstruction performance among approaches that produce models of similar size, such as PUP-3DGS~\cite{hanson2025pup} and Speedy-Splat~\cite{hanson2025speedy}. It slightly improves PSNR values in the Deep Blending~\cite{hedman2018deep} and Tanks \& Temples~\cite{knapitsch2017tanks} datasets. Our method achieves a favorable balance between quality, efficiency, and compactness across different datasets, maintaining competitive PSNR and SSIM scores while significantly reducing the size of the model. This highlights the effectiveness of our design in achieving compact 3D Gaussian representations with minimal visual quality degradation.

\vspace{2mm}\noindent\textbf{Qualitative comparison.}
    The subjective comparison between the baselines and our method is shown in \autoref{fig:Result}. Examples include the \textbf{Train} from Tanks and Temples~\cite{knapitsch2017tanks}, \textbf{Playroom} scenes from Deep Blending~\cite{hedman2018deep}, and \textbf{Treehill} scenes from Mip-NeRF 360~\cite{barron2022mip}. These results demonstrate that our method can reconstruct fine details, showing its effectiveness and adaptability across various scenes. 

\begin{figure}[t]
  \centering
   \includegraphics[width=1\linewidth]{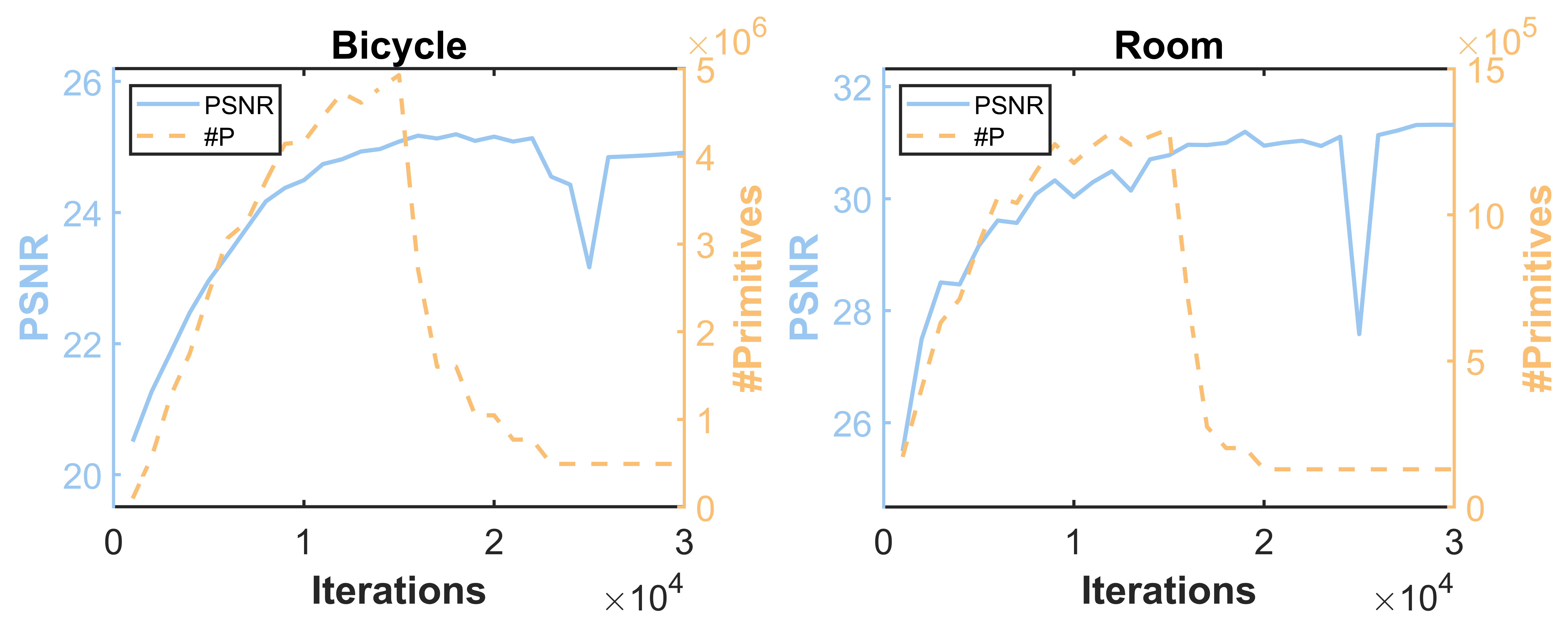}
   \caption{Variations in the primitive count and PSNR values of (Left) \textbf{Bicycle} and (Right) \textbf{Room} scenes using our method. The PSNR drop at the 25k iteration is due to the activation of 3D-DoG.}
\vspace{-0.3cm}
   \label{fig:curveresult}
\end{figure}

    \vspace{2mm}\noindent\textbf{Pruning analysis.} The pruning process and the PSNR increase curve are shown in~\autoref{fig:curveresult}. This figure illustrates that our Reconstruction-aware Pruning Scheduler can be successfully integrated into the original training pipeline, with progressive pruning maintaining a consistent upward trend in PSNR.  

    \autoref{fig:curveresult} also shows the trends in PSNR and primitive count during optimization. It demonstrates that our Reconstruction-aware Pruning Scheduler progressively compresses the overall number of primitives without hindering PSNR improvement. In some cases, such as the \textbf{Bicycle} scene (\autoref{fig:curveresult} left), we inevitably observe a slight performance drop in the late stage of pruning due to the aggressive 90\% pruning target set in our experiment. Note that the performance degradation in the 25k iteration is due to the activations of 3D-DoG attributes, which can be quickly recovered in subsequent iterations. 
\begin{figure*}[ht]
  \centering
  \vspace{-0.3cm}
   \includegraphics[width=1\linewidth]{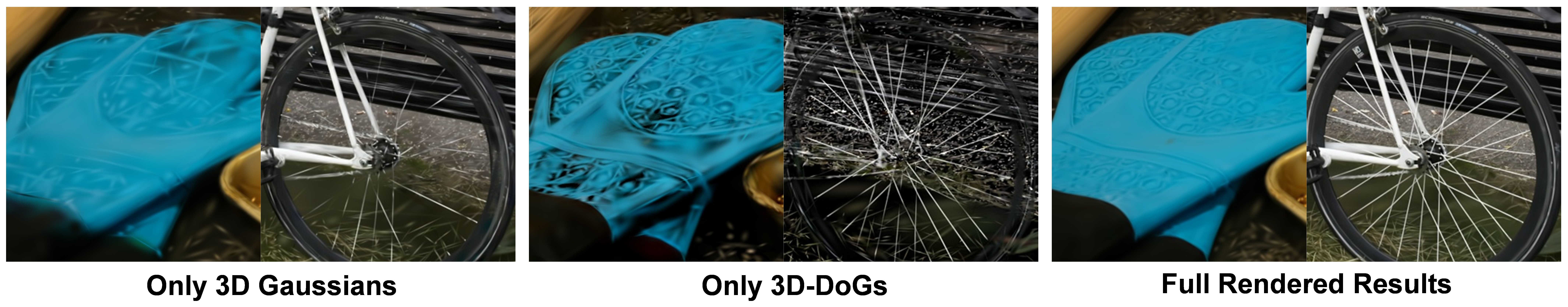}
   \caption{The rendering results are obtained by separating 3D Gaussians and 3D-DoGs in our 90\% primitives pruned 3DGS model. Compared to vanilla 3D Gaussian, the proposed 3D-DoG is more sensitive to capture details such as edges and textures. Examples are respectively from \textbf{Counter} and \textbf{Bicycle}.}
\vspace{-0.3cm}
   \label{fig:GorDoG}
\end{figure*}

   \vspace{2mm}\noindent\textbf{Collaboration Between 3D Gaussians and 3D-DoGs.} We isolate 3D Gaussians and 3D-DoGs in our mixture model for rendering, and the results are shown in~\autoref{fig:GorDoG}. These results demonstrate that the two primitives work collaboratively: 3D Gaussians capture the overall structure, while 3D-DoGs are predominantly placed around texture and edge regions.

\begin{figure}[t]
  \centering
   \includegraphics[width=1.0\linewidth]{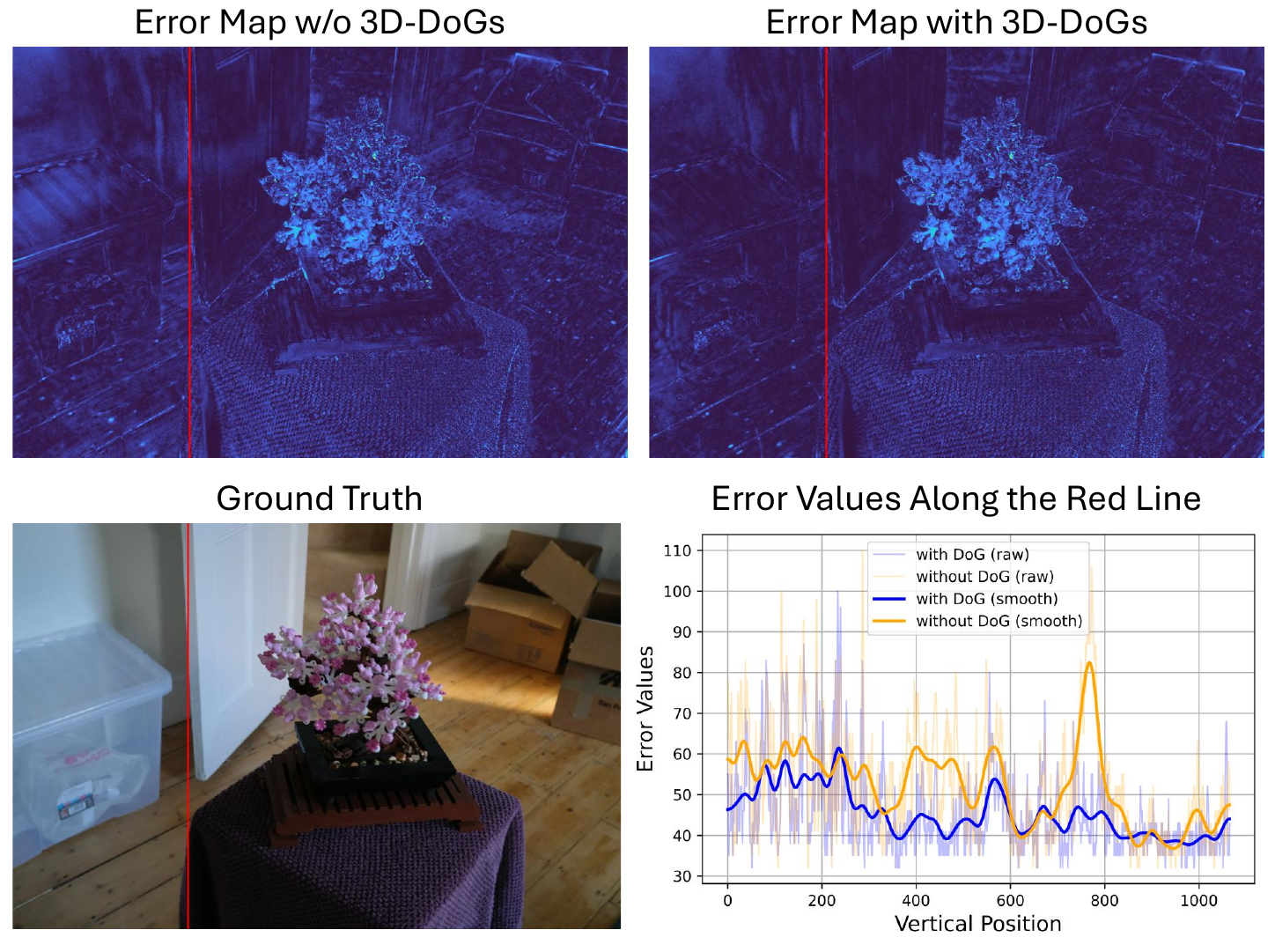}
   \vspace{-4mm}
   \caption{\textbf{Error Comparison.} (Top): Error heatmaps over the full image for the compact model without (left) and with (right) 3D-DoGs. (Bottom-left): Novel view of the \textbf{Bonsai} scene. (Bottom-right): Reconstruction error along the red line for both models.} 
   \label{fig:redline}
   \vspace{-4mm}
\end{figure}

    \autoref{fig:redline} compares the error maps of a compressed model without and with 3D-DoGs in the \textbf{Bonsai} scene along a red line that traverses both edge and textured regions. The error values along this line are generally lower when 3D-DoGs are incorporated, indicating that 3D-DoGs effectively reduce reconstruction errors, particularly at structural boundaries and in texture-rich areas.

\subsection{Ablation Study}
    We isolate our contributions using a set of modified frameworks, starting with the 3DGS backbone with 100\% primitives. We first use the refinement interval Regulation (RIR) as \textbf{V1} to prune 90\% of primitives. At each step, prune the same number of primitives based on opacity rank. In \textbf{V2}, the Dynamic Pruning Ratio Adjustment (DPRA) is integrated to tune the number of pruning primitives at each step. The Spatio-spectral Pruning Score (SPS) is adopted to replace opacity to measure per-primitive importance in the 3DGS model in \textbf{V3}. Finally, we introduce a novel 3D-DoG primitive into the architecture to formulate our method. 
    
    \autoref{tab:ablation_combined} provides a comprehensive comparison of both reconstruction quality and efficiency across different model variants. As the model scales, the variant with only 10\% of the parameters consistently approaches the full 100\% model in performance, validating the effectiveness and general robustness of our proposed architecture. In terms of efficiency, our techniques significantly accelerate 3DGS training and inference. Specifically, \textbf{V1} achieves approximately a $1.36\times$ reduction in training time and nearly $2.5\times$ higher FPS compared to the original 3DGS baseline. \textbf{V2} further improves training efficiency while maintaining a stable rendering speed. \textbf{V3} introduces SPS, slightly increasing train time without a noticeable loss in FPS. Finally, our complete model (“Ours”), which incorporates the 3D-DoG module, delivers the best reconstruction quality, with only a marginal drop in efficiency due to the additional computation introduced by 3D-DoG primitives but still accelerates the training by $1.23\times$ and inference by $2\times$ respectively. 

    \begin{table}
    \centering
    \small
    \setlength{\tabcolsep}{3pt}
    \captionsetup{singlelinecheck=false}
    \vspace{-0.3cm}
    \caption{Comparison of different variants in terms of reconstruction quality and efficiency on the Mip-NeRF360 dataset using 3DGS as the backbone. $\uparrow$ indicates higher is better, while $\downarrow$ indicates lower is better.}
    \resizebox{\columnwidth}{!}{
    \begin{tabular}{r|cccc | ccc | cc}
    \multicolumn{1}{l}{}  &\multicolumn{4}{c}{\textbf{Components}} & \multicolumn{3}{c}{\textbf{Reconstruction Metrics}} & \multicolumn{2}{c}{\textbf{Efficiency Metrics}} \\
    Variant & RIR  & DPRA  & SPS  & 3D-DoG & PSNR$\uparrow$ & SSIM$\uparrow$ & LPIPS$\downarrow$ & Time$\downarrow$ & FPS$\uparrow$  \\
    \hline
     3DGS &\xmark  & \xmark & \xmark &\xmark & 27.47 & 0.826 & 0.201 & 17m1s &  143.5 \\
     \hline
     V1 &\cmark  & \xmark & \xmark & \xmark & 26.03  & 0.742 & 0.331 & \cellcolor{lightorange}12m17s & \cellcolor{lightorange}362.4 \\
     V2 &\cmark  & \cmark & \xmark & \xmark & \cellcolor{lightyellow}26.17 & \cellcolor{lightyellow}0.751 & \cellcolor{lightyellow}0.324 & \cellcolor{lightred}11m34s & \cellcolor{lightred}363.2 \\
     V3 &\cmark  & \cmark & \cmark & \xmark & \cellcolor{lightorange}26.99  & \cellcolor{lightorange}0.771 & \cellcolor{lightorange}0.299 & \cellcolor{lightyellow}13m28s & \cellcolor{lightyellow}361.9 \\
     Ours &\cmark & \cmark & \cmark & \cmark & \cellcolor{lightred}27.16  & \cellcolor{lightred}0.789  & \cellcolor{lightred}0.285 & 13m48s & 289.0 \\
    \end{tabular}
    }
       \vspace{-2mm}
    \label{tab:ablation_combined}
\end{table}

\section{Conclusion}

We propose an adaptive framework for efficient 3D Gaussian Splatting that jointly optimizes pruning and reconstruction quality. By introducing a reconstruction-aware pruning scheduler, our method dynamically balances compression and quality during training, enabling stable convergence and compact representations. In addition, the proposed 3D Difference-of-Gaussian (3D-DoG) primitives enhance the expressive power of compact models, allowing them to retain fine structural details even under aggressive pruning. Through extensive experiments on multiple benchmarks, our approach consistently achieves a favorable trade-off between rendering fidelity, model size, and computational cost. The results suggest that adaptive, reconstruction-driven compression can be a promising direction for scalable 3D scene representation. Future work includes extending the framework to dynamic, large-scale scenes and integrating it with hardware-efficient rendering for real-time use.

{
    \small
    \bibliographystyle{ieeenat_fullname}
    \bibliography{main}
}

\end{document}